\pdfoutput=1

\documentclass[11pt,a4paper]{article}
\usepackage[hyperref]{acl2020}
\usepackage{times}
\usepackage{latexsym}

\usepackage{microtype}

\aclfinalcopy %

\newif\ifwithappendix\withappendixtrue

\newif\ifappendixshown
\appendixshowntrue

\usepackage[T1]{fontenc}
\usepackage[utf8]{inputenc}
\usepackage[textsize=tiny]{todonotes}
\usepackage{graphicx}
\usepackage{booktabs}
\usepackage{latexsym}

\usepackage{footnote}
\makesavenoteenv{tabular}
\makesavenoteenv{table}
\makesavenoteenv{table*}

\newcommand\minput[1]{%
  \input{#1}%
  \ifhmode\ifnum\lastnodetype=11 \unskip\fi\fi}

\usepackage{hyperref}
\usepackage{xstring}

\newcommand{\noqa}[1]{}
\newcommand{\noqall}[1]{}

\usepackage{amsmath}
\usepackage{recycle}
\title{On the Multi-Property Extraction and Beyond}

\author{
 Tomasz Dwojak \and Michał Pietruszka \and Łukasz Borchmann \\
 {\bf Filip Graliński} \and {\bf Jakub Chłędowski} \\
 \\
 Applica.ai \\
  {\tt firstname.surname@applica.ai} \\
}

\begin{document}
\maketitle
\begin{abstract}
In this paper, we investigate the Dual-source Transformer architecture on the \emph{WikiReading} information extraction and machine reading comprehension dataset.
The proposed model outperforms the current state-of-the-art by a~large margin. 
Next, we introduce \emph{WikiReading Recycled} --- a~newly developed public dataset, supporting the task of multiple property extraction.
It keeps the spirit of the original \emph{WikiReading} but does not inherit identified disadvantages of its predecessor.

\end{abstract}

\def \OldWikiReading {\textsc{WikiReading}}%
\def \NewWikiReading {\textsc{WikiReading Recycled}}%

\section{Introduction}

The \OldWikiReading{} dataset proposed by~\citet{hewlett-etal-2016-wikireading} is built on top of Wikipedia articles and properties taken from the WikiData database~\cite{WikiData_article}.
Its objective is to determine property-value pairs for provided text, e.g. to extract or infer information regarding the described person's occupation, spouse, \textit{alma mater} or place of birth, given related biographic article.
An important part of the task is to create a~model that is able to extract properties that have not appeared during training. 

Our approach to the aforementioned dataset relies on the Transformer architecture modified in order to support two source sequences~\cite{NIPS2017_7181, junczys-dowmunt-grundkiewicz-2018-ms}.
The proposed model consists of a~single decoder that generates property values and of two encoders with shared weights: one for property names and one for the article to analyze.

Our work on \OldWikiReading{} inspired us to create the \NewWikiReading{} dataset to extract multiple properties of the same object at once.
The dataset uses the same data as the \OldWikiReading{} but unlike in the original dataset, validation and test sets do not share articles from the train set. Additionally, the test set contains properties not seen during training, posing a~challenging subset for current state-of-the-art systems.
The provided human-evaluated test set contains only those properties that could be inferred from the article.
Finally, a~strong Dual-source Transformer baseline for the \NewWikiReading{} task is provided.

\section{Related Work}
Early work in relation extraction revolves around problems crafted using distant supervision methods~\cite{Craven:1999:CBK:645634.663209}. %
Encoder-decoder models, that had been previously framed in NMT problems~\cite{bahdanau2014neural}, have recently been used in solving information extraction problems formulated with triples (property name, property value, item)~\cite{vu2016combining}, as well as in similar problems of Question Answering~\cite{DBLP:journals/corr/FengXGWZ15}. The difference between \OldWikiReading{} and QA problems is in how questions are being asked, namely whether they are formulated in natural language or as a~raw property name.

In response to this popular discourse, a~\OldWikiReading\ dataset with millions of training instances was proposed~\cite{hewlett-etal-2016-wikireading}.
Many baseline methods were evaluated alongside the dataset. The best performing model (\textit{Placeholder seq2seq}) uses placeholders to allow rewriting out-of-vocabulary words to the output, achieving {\textsc{Mean-$F_1$}} score of $71.8$.

The following work of~\citet{choi-etal-2017-coarse} re-evaluated the \textit{Placeholder seq2seq} model and reached a~{\textsc{Mean-$F_1$}} score of $75.6$. Moreover, the authors proposed a~reinforcement learning approach which improved results on the~challenging subset of 10\% longest articles. %
This framework was extended by~\citet{MSCQA_System} with the addition of a~self-correcting action, that removes the inaccurate answer from the GRU-based~\cite{chung2014empirical} answer generation module and continues to read, reaching a~75.8 {\textsc{Mean-$F_1$}} score on the whole \OldWikiReading.

\citet{hewlett-etal-2017-accurate} holds the state-of-the-art on \OldWikiReading\ with their proposition of SWEAR --- a~hierarchical approach that attends over a sliding window's GRU-generated representations in order to reduce documents to one vector from which another GRU network generates the answer. Additionally, authors set up a~strong semi-supervised solution on a~1\% subset of \OldWikiReading.

\section{Dual-source Transformer\label{sec:transformer}}

The Transformer architecture proposed by~\citet{NIPS2017_7181} was further extended to support two inputs by~\citet{junczys-dowmunt-grundkiewicz-2018-ms} and successfully utilized in Automatic Post-Editing. We propose to apply this Dual-source Transformer in information extraction and machine reading comprehension tasks.

The architecture consists of two encoders that share parameters and a~single decoder. Moreover, both the encoders and decoder share embeddings and vocabulary. In our approach, the first encoder is fed with the text of an article, and the second one takes the names of properties to determine. %

Datasets were processed with a~SentencePiece model~\cite{kudo-2018-subword} trained on a~concatenated corpus of inputs and outputs with a~vocabulary size of 32,000. Dynamic batching was applied during training, in order to use the GPU memory optimally (nevertheless, the average batch size was around 100). The model was implemented in Marian NMT Toolkit~\cite{mariannmt} and its specification followed the default Marian's settings for Transformer models. The only difference was reduction of encoder and decoder depths to 4.\footnote{The complete configuration file will be available on GitHub.}

\begin{figure}
    \centering
    \includegraphics[width=0.48\textwidth]{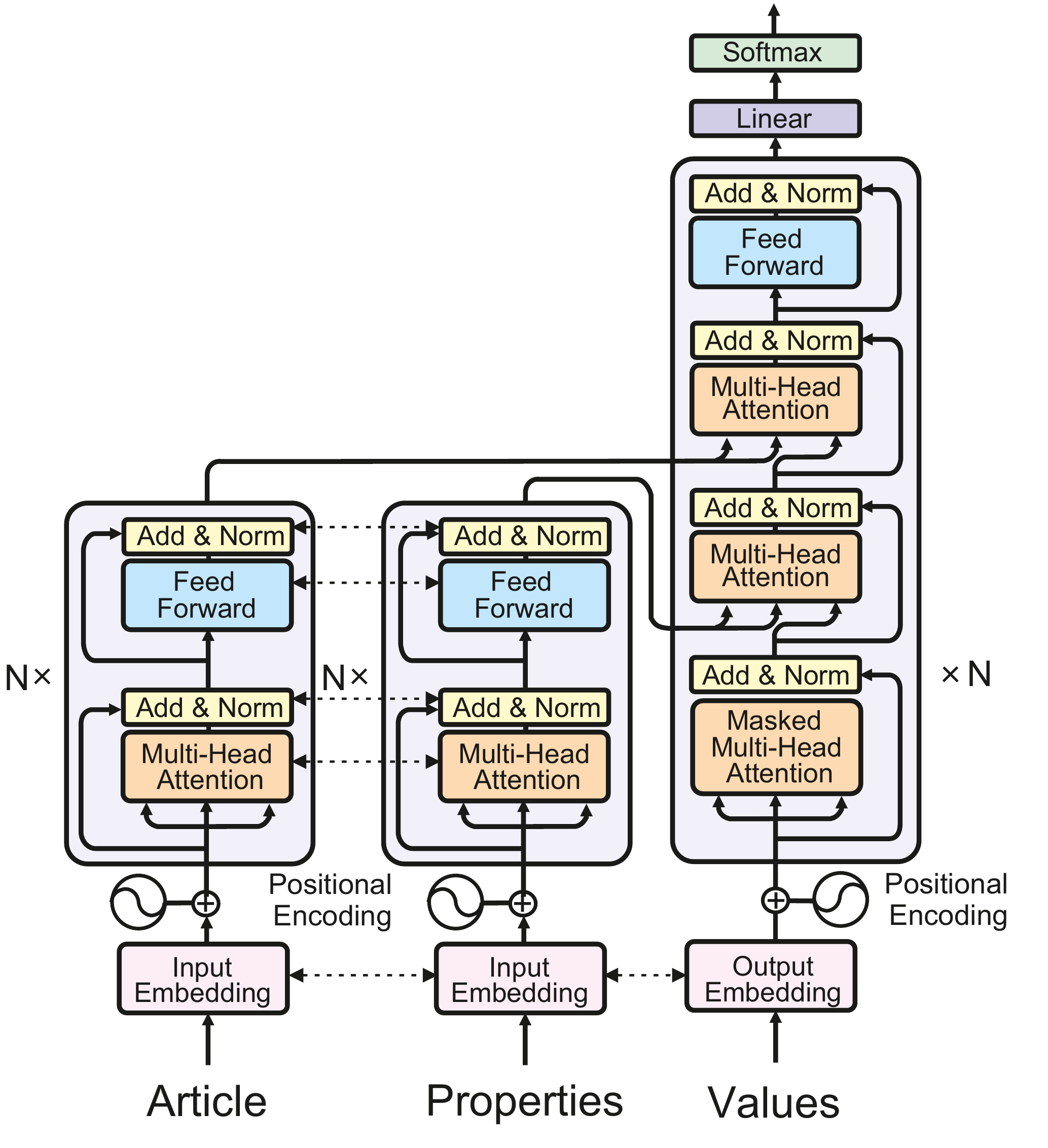}
    \caption{Architecture of Dual-source Transformer as proposed by~\citet{junczys-dowmunt-grundkiewicz-2018-ms} for Automatic Post-Editing. In the case of \NewWikiReading{} and \OldWikiReading{}, the encoder transforms an article and the corresponding properties separately.}\label{fig:my_label}
\end{figure}{}

\section{ \NewWikiReading}

\NewWikiReading\ introduces the problem of multi-property information extraction with the goal of evaluating systems that extract any number of given properties at once from the same source text.
It is built on \OldWikiReading, the biggest publicly available dataset for information extraction, with improved design and human annotation.
In order to achieve that, we propose to merge data instances from all splits (training, validation, and test sets) that contain the same articles by combining their property names and values.
The resulting dataset contains approximately 4.1M instances with 703 distinct properties that we split into new training, validation, and test sets.

We perceive the model's generalization abilities (i.e. to extract unseen properties) as an important metric. Therefore, we assigned 20\% of the properties to the test set only. In order to make the validation set a~good approximation of the test set, another 20\% of the properties are validation-only and a~set of 10\% of the properties are shared between the test and validation splits. This leads to a~design where as much as 50\% of the properties cannot be seen in the training split, while the remaining 50\% of the properties can appear in any split.%

The quality of test sets plays a~pivotal role in reasoning about systems performance. Therefore, a~group of annotators went through the instances of the test set and assessed whether the properties either appear in the article or can be inferred from it.
Relevance of the aforementioned validation can be demonstrated by the fact that {\textsc{Mean-$F_1$}} on a~pre-verified set was approximately $0.2$ lower and 8\% of articles were removed completely. Apparently, 28\% of property values were marked as unanswerable and were removed.

This led to the creation of a new test set, where the proportion of the properties has slightly changed. Hence, 27\% of the properties in the test set were not seen during the training and 15\% are test set only. Similarly, 36\% of the validation set has not been seen during training.

It was determined that 46\% of expected values in the test set were present in the article explicitly, whereas 54\% of test set values were possible to infer.  %

\begin{table}
\centering
\begin{tabular}{@{}lrrr@{}}
\toprule
Data split & Total & Overlap & \% \\ \midrule
validation set &  1,452,591 & 1,374,820 & 94.65 \\
test set &  821,409 & 780,639 & 95.04 \\ \bottomrule
\end{tabular}
\caption{The size of \OldWikiReading\ splits. Column (\emph{Total}) shows the total number of unique Wikipedia articles in each split and column (\emph{overlap}) shows the number of articles that have been seen in the train set. The last column shows the percentage of the overlap between the considered set and train set. }
\label{tab:1}
\end{table}

\section{Evaluation}

\def \WROrig {\textsc{Mean-$F_1$}}
\def \OurMeanF {\textsc{Mean-MultiLabel-$F_1$}}
Performance of systems is evaluated using the F1 metric, adapted for the \NewWikiReading\ specifics. For the \OldWikiReading\ metric, \WROrig\ follows the originally proposed metric and assesses F1 scores for each property instance, to be averaged later over the whole test split. We extend this metric due to changes introduced in the instance definition with the new metric \OurMeanF\, which is able to handle multiple properties, where each can have multiple answers. The \OurMeanF\ score is calculated for each property name and then averaged per article, and in the next step averaged over all articles. \OurMeanF\ is invariant to the order of generated answers.

It is worth noting that the \NewWikiReading\ instances could contain multiple property names. Due to that, models trained on \NewWikiReading\ are able to use solely the context of one property to deduce correct answers about another properties. As a result, a model trained on property names achieves up to a~0.18 \OurMeanF score without seeing the actual article content. One such example would be answering the property \textit{instance of: human} just by seeing another property name \textit{educated at}.

\subsection{Baseline\label{sec:baseline}}
To compare with the previous results, we reproduce a basic sequence to sequence model from~\citet{hewlett-etal-2016-wikireading}.
Since the model's description missed some important details, they had to be assumed before model training. We supposed that the model consisted of unidirectional LSTMs and it was trained with mean (per word) cross entropy loss, until no progress was observed for 10 consecutive validations occurring every 10,000 updates. Input and output sequences were tokenized and lowercased. In addition, truecasing was applied to the output. We use syntok\footnote{\url{https://github.com/fnl/syntok}} tokenizer and a~simple RNN-based truecaser proposed by~\citet{susanto-etal-2016-learning} were used. During inference, we used a~beam size of 8. The rest of parameters followed the description provided by the authors\footnote{The complete configuration file will be available on GitHub.}.

\subsection{Results on \OldWikiReading{}}

The reproduced \textit{Basic seq2seq} model achieved a {\textsc{Mean-$F_1$}} score of $0.748$, that is 3 points higher than reported by~\citet{hewlett-etal-2016-wikireading} and less than 1 point lower than \textit{Placeholder seq2seq} reimplemented by~\citet{choi-etal-2017-coarse}.
Hence, the Results of our reimplementation may suggest that the method proposed in the initial \OldWikiReading{} paper suffered due to poor optimization.

We evaluated two training approaches for the dual-encoder model.
In the first scenario, we merge all property names related to the given article (Multi-property).
In the second one, we train the model on each property name separately (Single-property).
Nonetheless, the evaluation process was performed in a single-property manner.

The dual-encoder solution outperforms previous state-of-the-art models. The single-property model achieves a~slightly higher performance of 79.9\%.
\begin{table}
\centering

\begin{tabular}{@{}lr@{}}
\toprule
Model & {\textsc{Mean-$F_1$}}\\ \midrule
Basic s2s &~74.8 \\
Placeholder s2s~\cite{choi-etal-2017-coarse} &~75.6 \\
\vspace{5pt}
SWEAR~\cite{hewlett-etal-2017-accurate} &~76.8 \\
Dual-source Transformer &  \\

\quad Multi-property & 79.4 \\
\quad Single-property &  \textbf{79.9} \\
\bottomrule
\end{tabular}
\caption{Results on \OldWikiReading{} (test set). \textit{Basic s2s} denotes the re-implemented model described in Section~\ref{sec:baseline}.}
\label{tab:2}
\end{table}

\subsection{Results on \NewWikiReading{}}
Finally, we propose two models as baselines for \NewWikiReading{}: the reproduced \textit{Basic seq2seq} and the dual-encoder model.
In addition, we evaluate an ensemble of four best-performing checkpoints on the validation set.
Table~\ref{tab:3} presents \OurMeanF scores on the test set.
The dual-encoder model outperforms the \textit{Basic seq2seq} as in the case of \OldWikiReading\ task.
The former achieved {\textsc{Mean-MultiLabel-$F_1$}} of 79.5\%.
Additionally, the ensemble submission improved the single-best model by 0.5 points.

Moreover, test set was split into two subsets for analytic purposes.
The first resulting subset contains property values that appear in the article explicitly (exact-matches, EM), whereas the second contains the rest of data, e.g. the property values that are inferable (IN).
Since the precise computation of precision is impossible in this scenario (one cannot determine which incorrect values were predicted for which expected ones), we report only recall on these subsets.
The single-best model achieves the highest scores on both subsets: 73.3\% on the exact-match subset and 73.9\% on the inferable one.

Additionally, we evaluate both models on the subset of property names that did not appear in the training set. To our surprise, both models perform poorly. The \textit{Basic seq2seq} model achieves 2.4\% (\OurMeanF{}), whereas the dual-source model ignores those properties and does not generate answers for them.

\begin{table}
\centering
\begin{tabular}{@{}lrrr@{}}
\toprule
Model                   & MM-$F_1$                       & EM    & IN    \\ \midrule
Basic s2s               & 75.2                           & 66.3  & 70.8  \\
Dual-source Transformer & \\
\quad Single            & 79.5          & \textbf{73.3}  & \textbf{73.9}  \\
\quad Ensemble          & \textbf{80.0} & 73.1           & 73.8\\
\bottomrule
\end{tabular}
\caption{Results on \NewWikiReading{}. We chose the model with the highest score on the validation set for the final submission (single) and ensemble of the four best-performing checkpoints. MM-$F_1$ stands for \OurMeanF, EM for exact-match subset, and IN for inferable property subset. Note that subsets were evaluated with recall instead of MM-$F_1$.}%
\label{tab:3}
\end{table}

\section{Summary}
We showed that the Dual-source Transformer outperforms the previous state-of-the-art model on the \OldWikiReading{} by a~far margin. The architecture was successfully adapted from Automatic Post-Editing systems to information extraction and machine reading comprehension tasks.

Moreover, \NewWikiReading{} was introduced --- to the best of our knowledge, the first multi-property information extraction dataset with a~human-annotated test set. In this case, a~different setting of Dual-source Transformer was applied, significantly outperforming the presented baseline approach.

Both the dataset and models, as well as their detailed configurations required for reproducibility, have been made publicly available.

An analysis of our results on a~challenging subset of unseen properties reveals that despite high overall scores, existing systems fail to provide satisfactory performance. Low scores indicate an opportunity to improve, as these properties were verified by annotators and are expected to be answerable. We look forward to seeing models closing this gap and leading to remarkable progress in the field of machine reading comprehension.

\bibliography{ms}
\bibliographystyle{acl_natbib}

\end{document}


\newpage
\appendix
\section{Appendices}
\label{sec:appendix}

Instances from the original WikiReading dataset were merged to produce 4,124,422 instances of a~new `per article' type and split with the assumption that a~given proportion of 20\% labels (label set 1) should be unique in the test set(block A), more precisely not seen before in train and validation, thus allowing to measure systems ability to generalize to unseen relations. To achieve this, 1k articles were drafted in such a~way that its properties contains types not seen in the remaining subsets. In order to make validation a~good estimation of a~test set without disclosing particular labels, unique validation-only labels were introduced (block B). Again, 1k articles with 20\% of labels (label set 2) not seen in the remaining subsets were drafted. Additionally, test and validation set shares 10\% of labels  (label set 3) that does not occur in the train set adding 2k of articles to each of these subsets(block C, D). The remained 4,118,422 articles has labels strictly from the remaining 50\% of the label set (label set 3), of which 2k examples were added into each test (block E) and validation set (block F). From the rest we build a~training set consisting of 4,114,422 examples (block G).

It is worth noting that an example from block A~that has some unique property names, can also posses properties from blocks C-G, and an example from block B can have properties.

The table \ref{table:1} is a~summary of referenced blocks.
 \begin{table}[h!]
\centering
\begin{tabular}{||c c c c c c||}\noqa
\hline
\multicolumn{4}{|c|}{Label set} \\
 Block & set 1 & set 2 & set 3 & set 4  \\ [0.5ex]
 \hline\hline
 A & must have & not allowed & allowed & allowed \\
 B & not possible & must have & allowed & allowed \\
 C & not possible & not possible & must have & allowed \\
 D & not possible & not possible & must have & allowed \\
 E & not possible & not possible & not possible & must have \\
 F & not possible & not possible & not possible & must have \\
 G & not possible & not possible & not possible & must have \\
 [1ex]
 \hline
\end{tabular}
\caption{The table shows relation between label sets and blocks of articles. `must have' relation denotes that a~given articles must have any label from the corresponding set. `Not Allowed' labels were removed from a~particular block of articles, whereas `allowed' does not pose any limitations to its usage. `not possible' means that labels presence in a~corresponding block cannot happen in principle due to the splitting algorithm design.}
\label{table:1}
\end{table}

\section{Appendices}
\label{sec:appendix}
\begin{itemize}
    \item art --- article
    \item key --- property name
    \item value --- property value
\end{itemize}

\subsection{}
Metric used in the original WikiReading\\
$
\displaystyle Mean F1(keys, out\_values, expected\_values ) = \frac{1}{\#keys}\sum_{k=1}^{\#keys}F1(out\_values(k), expected\_values(k))
$
\\
\\

Proposed metric used in the new WikiReading (works for one\_article- multiple\_properties\_settings)\\
$
\displaystyle Mean Multilabel F1(articles, keys, out\_values, expected\_values ) = \frac{1}{\#articles} \sum_{a=1}^{\#articles}  \frac{1}{\#keys} \sum_{i=1}^{\#keys} F1(out\_values(a,k), expected\_values(a,k))
$
\\
\\

Per label metric (useful when showing how does a~system generalize to unseen labels)\\
$
\displaystyle Multilabel F1(articles, key, out\_values, expected\_values ) = \frac{1}{\#articles} \sum_{a=1}^{\#articles} F1(out\_values(a, key), expected\_values(a, key))
$.